# Leveraging Federated Learning and Edge Computing for Recommendation Systems within Cloud Computing Networks

Yaqian Qi [1]*
Quantitative Methods and Modeling
Baruch Collegue, CUNY
55 Lexington Ave, New York, NY 10010
* Corresponding author: alicia.qi.yaqian@gmail.com

Yuan Feng [1]
Interdisciplinary Data Science
Duke University
North Carolina USA
yuan.feng.dsduke@gmail.com

Xiangxiang Wang [2]
Computer Science
University of Texas at Arlington
Arlington, Texas
wx18714999@gmail.com

Hanzhe Li [3]
Computer Engineering
New York University
New York USA
Nyhanzheli@gmail.com

Jingxiao Tian [4]
Electrical and Computer Engineering
San Diego State University
San Diego, US
jtian1125@sdsu.edu

*Abstract*—To enable large-scale and efficient deployment of artificial intelligence (AI), the combination of AI and edge computing has spawned Edge Intelligence, which leverages the computing and communication capabilities of end devices and edge servers to process data closer to where it is generated. A key technology for edge intelligence is the privacy-protecting machine learning paradigm known as Federated Learning (FL), which enables data owners to train models without having to transfer raw data to third-party servers. However, FL networks are expected to involve thousands of heterogeneous distributed devices. As a result, communication efficiency remains a key bottleneck. To reduce node failures and device exits, a Hierarchical Federated Learning (HFL) framework is proposed, where a designated cluster leader supports the data owner through intermediate model aggregation. Therefore, based on the improvement of edge server resource utilization, this paper can effectively make up for the limitation of cache capacity. In order to mitigate the impact of soft clicks on the quality of user experience (QoE), the authors model the user QoE as a comprehensive system cost. To solve the formulaic problem, the authors propose a decentralized caching algorithm with federated deep reinforcement learning (DRL) and federated learning (FL), where multiple agents learn and make decisions independently.

## I. INTRODUCTION

Linhua Technology, the world's leading edge computing solution provider, and to Star Technology (referred to as "Nebula Clustar") have entered into a cooperation to deal with the data delay and privacy protection issues in centralized machine learning training, and work together to create an all-in-one machine for edge federated learning. This product uses MECS-7211 of Linhua Technology as the edge computing server, and the FPGA privacy computing acceleration card of Nebulstar, fully integrates the advantages of high-performance privacy computing and edge computing, and effectively meets the business requirements of intensive edge computing scenarios. Such as distributed machine learning, gene sequencing, financial business, medical, video processing, network security and so on.

At present, with the rapid development of the Internet of Things and the popularity of 5G network, a large number of terminal devices are connected to the network to generate massive data. Traditional data calculation and analysis are based on cloud computing, but with the rapid increase of data, the process of transmitting from application terminals to cloud computing centers will cause delay and data leakage. How to process data in a timely and efficient manner has become a major challenge for cloud computing centers. Edge Computing, a new computing model for computing at the edge of the network, emerged. At the edge of the network near people, things or data sources, edge intelligent services are provided nearby to respond with more efficient network services to meet the increasing needs of many businesses such as the Internet of Things, the Internet of vehicles, industrial control, intelligent manufacturing, and large video.

On the other hand, the introduction of edge computing technology has reduced the network burden of cloud centers, but it has also caused security problems, and the localization of data is easy to hinder the interaction between data, coupled with the continuous tightening of data security and application specifications in recent years, such as GDPR data privacy and data protection issues have been highly valued. The centralized calculation of data adopted by traditional machine learning algorithms cannot cope with the requirements of data specifications, which limits the development of artificial intelligence.

In this context, Federated machine learning (Federated machine Learning) should be born in time to provide a solution to the security problem of edge computing. Federated learning is a machine learning framework that only exchanges the parameters, weights and gradients of the model after encryption, without moving the original data out of the local area or moving the encrypted original data into a centralized location, which can help multiple organizations to meet the requirements of user privacy protection, data security and government regulations. Therefore, in order to improve caching benefits, this paper proposes an edge cache and recommendation system, which forms an edge cache system supporting recommendation in mobile edge cloud computing network. The proposed system supports both direct hit and soft hit. The factors that affect the user's QoE are modeled as the comprehensive system cost (including similarity cost, delay cost and cache hit cost). The authors further formulates the cache replacement problem as a multi-agent Markov decision process (MDP) to minimize the expected long-term system cost (reflecting the user QoE).

## II. RELATED WORK

### A. The combination of edge computing and federated learning

The combination of edge computing and federated learning is designed to take advantage of edge computing's advantages in data processing speed and response time, while protecting user data privacy through the distributed model training approach of federated learning. This combination has shown its potential in several areas, especially in recommendation systems, which are able to provide personalized recommendations while protecting user privacy.

Federated Learning (FL) is a distributed machine learning framework that allows users to train models using local data sets. For participating nodes, they need to have sufficient computing power, network bandwidth and storage capacity. But in federal edge learning (FEL), edge device resources are relatively limited. These devices need to operate for a long time and are capable of constantly updating their models. They participate in the training process only when they are free and can withdraw from the process at any time. As a result, the overall system needs to be more flexible, scalable, and support low-overhead long-term model training and updating requirements to meet the constraints and requirements of edge devices.

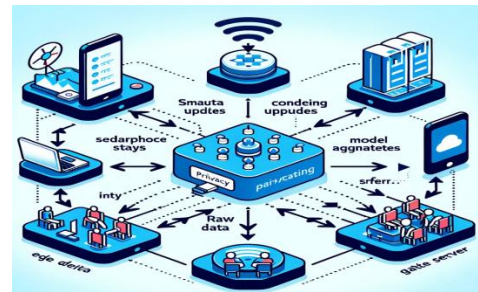

Figure 1. Simplified federated learning architecture diagram

The simplified illustration of a Federated Learning architecture is now displayed, showing a more straightforward view of the concept with a few edge devices connected to a central server. This design emphasizes the privacy-centric approach of Federated Learning, where only model updates are communicated to the server, ensuring that raw data remains on the local devices.

At the same time, although federated learning avoids the direct exchange of sensitive data between participants, the exchange of plaintext parameters still brings the risk of privacy disclosure. Many existing privacy protection schemes used in federated learning can ensure the security of transmission parameters, but many privacy protection schemes may bring a lot of extra computing and communication costs, and it is difficult to balance the relationship between privacy protection and efficiency.

### B. Recommendation system based on federated learning

Similar to the architecture design in the general federated learning domain, the architecture used in the research of federated recommendation systems can be divided into client-server architecture and decentralized architecture, as shown in Figure 2.

(1) The training process of client-server architecture is as follows:

- The server initializes the model parameters and sends the model parameters to each client.
- The client uses the local data and the latest model parameters received from the server for training, and sends the intermediate parameters to the server;
- The server aggregates the intermediate parameters, updates the global model, and sends the model back to the client.

Repeat steps (2) and (3) until the model converges.

Features: This type of architecture can use computing resources on the server side and reduce the computing pressure on the client side, but it is prone to single point of failure. At the same time, for the curious server, it may infer the privacy information of the client according to the intermediate parameters uploaded by the client, thus revealing the privacy of the client.

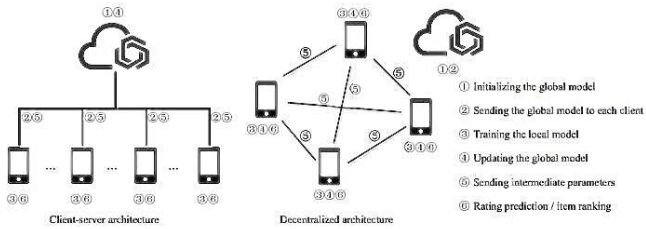

Figure 2. Federated Learning intelligent recommendation architecture design

(2) The training process of decentralized architecture is as follows:

The server initializes the model parameters and then sends the model parameters to each client.

The client uses local data to train the model and update the local model parameters.

The client selects some other clients, sends the local intermediate parameters, receives the intermediate parameters of other clients, and updates the local model;

Repeat steps (2) and (3) until the model converges.

Features:

- Anonymity. In the process of model training, the client can anonymously send intermediate parameters to other clients, so as to solve the problem of privacy disclosure between clients.
- Save server resources. The server only needs to initialize the model parameters and distribute the model parameters to each client, and does not need to participate in the update of the model.
- High availability. There is no single point of failure, that is, the failure of a single component of the server will not cause the entire federated learning system to stop training.

The similarities between the two: the original data of the client does not leave the local, and a shared model is obtained by sending intermediate parameters through the communication between the server and the client or the c

## C. Federalization of the recommendation system

The federalization of recommendation models has some commonality, and the training framework of a federated recommendation model is usually suitable for other models with the same training mode. However, considering the different levels of privacy protection in different scenarios and the different privacy issues that may be brought by different model parameters, there are certain differences in the federalization process of different recommendation models. The federalization of models can be divided into three categories: the federalization of recommendation algorithms based on collaborative filtering, the federalization of recommendation algorithms based on deep learning and the federalization of recommendation algorithms based on meta-learning.

## D. Advanced Technologies and Challenges in Privacy Protection for Federated Learning and Edge Computing

The integration of federated learning (FL) with edge computing has heralded a new era in privacy-preserving artificial intelligence (AI), allowing data processing closer to the source and minimizing privacy risks. However, deploying these technologies at scale involves navigating a complex landscape of advanced privacy-preserving techniques and inherent challenges.

- Advanced Technologies Overview

Secure Multi-Party Computation (SMPC): SMPC enables parties to jointly compute a function over their inputs while keeping those inputs private. Its application in FL can protect data during aggregation, but it demands significant computational and communication resources, posing a challenge for edge devices.

Homomorphic Encryption (HE): HE allows computations to be performed on encrypted data, providing results that, when decrypted, match the results of operations performed on the plaintext. While HE offers strong privacy guarantees, its high computational overhead limits its practicality for resource-constrained edge devices.

Differential Privacy (DP): DP introduces randomness to the data or the model outputs, ensuring that individual data contributions are masked. Implementing DP in FL helps mitigate the risk of information leakage but balancing privacy with data utility remains a critical challenge.

- Practical Applications

Several industries are pioneering the application of these technologies, from healthcare, where patient data sensitivity is paramount, to finance and smart manufacturing, where operational efficiency and data privacy must coexist. These applications demonstrate the potential of FL and edge computing to revolutionize privacy-preserving data analysis and decision-making.

- Inherent Challenges

Computational and Communication Costs: The primary bottleneck is the high cost of implementing privacy-preserving algorithms on edge devices, which often have limited computational power and energy resources.

Technology Integration and Compatibility: Combining different privacy-preserving technologies in a cohesive manner that aligns with the decentralized nature of FL and edge computing requires innovative architectural solutions.

Regulatory Compliance and Standards: Navigating the global landscape of data protection regulations, such as GDPR, poses challenges for the deployment of FL and edge computing solutions across borders.

- Future Research Directions

Addressing these challenges calls for ongoing research into optimizing privacy-preserving algorithms for efficiency, developing lightweight versions suitable for edge devices, and creating flexible frameworks that can adapt to various regulatory environments. Moreover, exploring AI and machine learning techniques to enhance the efficiency of these technologies presents a promising avenue for future work.

By advancing our understanding and capabilities in these areas, we can unlock the full potential of federated learning and edge computing as cornerstones of a privacy-preserving digital future.

## III. SYSTEM DESCRIPTION AND FUNCTION DEFINITION

### A. Personalized federated learning

Consider a decentralized federated learning system with N edge devices. Each device trains a personalized model x; ∈ Rd(d is the dimension of model xi). Thus, the local objective function of device i is defined as follows:

$$R_i(x_i) = f_i(x_i) + \lambda h_i(x_i) \quad (1)$$

Where fi(xi) is the local loss function of device i, hi(xi) represents the regularization function, and in represents the regularization factor. In this article, we use model regularization to facilitate collaboration between each device and its neighbors. For device i, its regularization function can be formalized as:

$$h_i(x_i) = \frac{1}{|S_i|} \sum_{j \in S_j} ||x_i - x_j||^2 \quad (2)$$

Si indicates the neighbor set of device i, and ‖ · ‖ indicates L2 normal pattern.

### B. Combined with model pruning PFL

In this paper, we use adaptive model pruning to improve the communication efficiency of equipment. For device i, the following is the model formula function: $m_i \in \{0,1\}^d$ If $m_i(k)=1$, it means that the KTH neuron of the local model xi is preserved. If $m_i(k)=0$, the KTH neuron of xi is discarded. Thus, after pruning the model can be pressed $x_i = \tilde{x}_i \circ m_i$. In addition, due to model pruning, the regularization function of device i can be redefined as:

$$h_i(\tilde{x}_i) = \frac{1}{|S_i|} \sum_{j \in S_i} J(\|m_{i,j} \circ (\tilde{x}_i - \tilde{x}_j)\|^2) \quad (3)$$

$$J(\|m_{i,j} \circ (\tilde{x}_i - \tilde{x}_j)\|^2) = 1 - e^{-\|m_{i,j} \circ (\tilde{x}_i - \tilde{x}_j)\|^2} \quad (4)$$

Is a commonly used negative exponential function that $x_{i*j} = x_i \circ m_i$ represents the overlap between the device and the device's model mask.

In this article, we aim to minimize the loss function on the local data set:

$$F(X) = \frac{1}{N} \sum_{i=1}^{N} f_i(x_i) \quad (5)$$

And regularization function:

$$H(X) = \frac{1}{N} \sum_{i=1}^{N} h_i(x_i) \quad (6)$$

Thus, in a PFL considering model pruning, the optimization objective is defined as:

$$\min_{\tilde{X}} \left\{ \mathfrak{R}(\tilde{X}) \triangleq F(\tilde{X}) + \lambda H(\tilde{X}) \right\} \quad (7)$$

$$\tilde{X} = [\tilde{x}_1, \tilde{x}_2, \cdots, \tilde{x}_N] \quad (8)$$

Represents the model parameter matrix for all device combinations.

### C. The process of proposing the method

To minimize this, we propose a method that combines adaptive model pruning and neighbor selection, as shown in Figure 1. The proposed method mainly includes two steps:

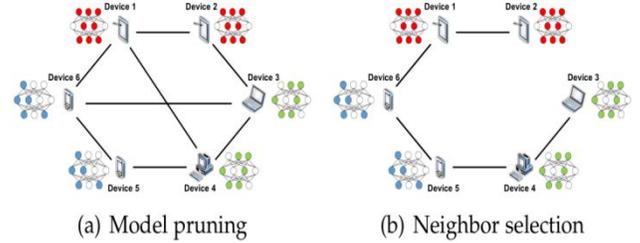

(a) Model pruning      (b) Neighbor selection

Figure 3. A method of joint model pruning and neighbor selection is proposed

## IV. EXPERIMENTAL EVALUATION

This paper introduces our novel algorithm, DPMN, and validates its effectiveness through experiments conducted on four widely recognized datasets:

### A. Experimental data

- Fashion-MNIST: Comprising 70,000 grayscale images distributed across 10 categories, each category contains 7,000 images. Specifically, the dataset is split into a training set with 6,000 images per category and a test set with 1,000 images per category. We employed the LeNet5 model for training on this dataset.

- CIFAR-10: This dataset includes 60,000 color images, categorized into 10 classes with each class having 6,000 images. It is divided into a training set with 5,000 images per class and a test set with 1,000 images per class. The LeNet5 model was also used for training on CIFAR-10.

- CEMNIST: Featuring a collection of 731,668 training samples and 82,587 test samples, the AlexNet model was utilized for training on this dataset.

- IMAGENET: This extensive dataset consists of 1,281,267 training samples, 50,000 validation samples, and 100,000 test samples, with each sample being a color image of size 224x224x3. Training on IMAGENET was conducted using the VGG16 model.

By employing these datasets, which vary significantly in size, complexity, and image type (ranging from grayscale to color images), we were able to thoroughly assess the performance and versatility of the DPMN algorithm across different visual recognition tasks.

B. *Experimental verification*

- DFL: Traditional decentralized training method.
- DPMP: A personalized model training method using model pruning
- DPCG: Neighbor selection method based on Euclidean distance for personalized model training
- Dpmn-r: Variant of DPMN, where neighbors are randomly selected

C. *Experimental result*

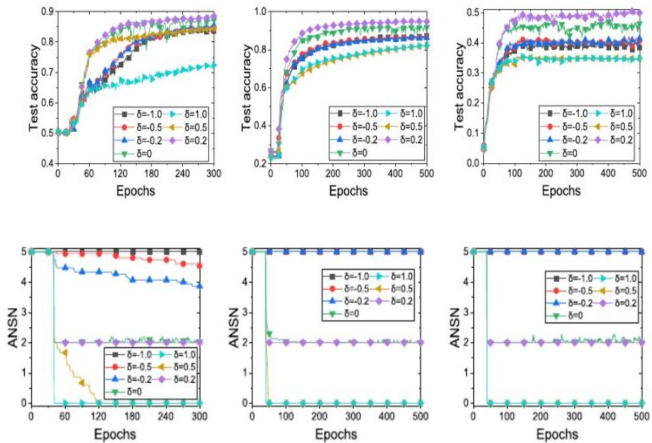

Figure 4. δImpact on DPMN performance

It can be seen from the experimental results that when the δ threshold of cosine similarity is set to 0.2, DPMN has the best effect and can achieve the balance between model quality and communication efficiency.

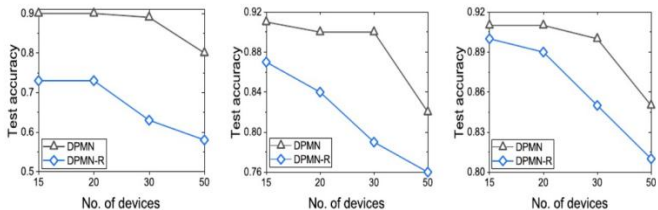

Figure 5. Accuracy vs. number of devices

From this group of experiments, it can be seen that in the same number of training rounds, our DPMN can significantly reduce the bandwidth resource consumption of model training, saving an average of 45.4% traffic overhead. When different number of devices participate in model training, the proposed device selection strategy can achieve higher model accuracy than the random neighbor selection method.

D. *Experimental conclusion*

The experimental evaluation of our novel algorithm, DPMN (Decentralized Personalized Model with Neighbor selection), underscores its significant potential in enhancing federated learning applications within the domain of recommendation systems, especially when integrated with edge computing strategies in cloud computing networks. The experiments conducted across diverse datasets—ranging from Fashion-MNIST and CIFAR-10 to CEMNIST and IMAGENET—demonstrate the versatility and efficiency of DPMN in handling various visual recognition tasks, which are pivotal in the context of personalized recommendation systems.

The comparison between DPMN and other decentralized training methods, including traditional decentralized learning (DFL), personalized model training with model pruning (DPMP), and neighbor selection based on Euclidean distance (DPCG), highlights the superiority of DPMN in balancing model quality with communication efficiency. Particularly, when the threshold of cosine similarity is set to 0.2, DPMN exhibits optimal performance, achieving a significant reduction in bandwidth resource consumption by an average of 45.4%. This efficiency is critical in cloud computing environments where bandwidth is a valuable resource.

Moreover, our findings reveal that the device selection strategy proposed by DPMN leads to higher model accuracy compared to random neighbor selection methods (Dpmn-r variant). This outcome is crucial for recommendation systems, where the accuracy of personalized recommendations directly impacts user satisfaction and engagement.

In conclusion, the application of federated learning leveraging edge computing, as embodied by our DPMN algorithm, presents a promising approach to improving recommendation systems within cloud computing networks. By efficiently managing communication overhead and enhancing model accuracy through personalized and context-aware learning, DPMN sets a new benchmark for the development of more responsive, efficient, and user-centric recommendation systems in the era of cloud and edge computing. This research not only validates the effectiveness of DPMN but also opens avenues for further exploration into optimizing federated learning frameworks for complex, data-driven applications in cloud computing ecosystems.

V. CONCLUSIONS

In this comprehensive study, we introduced the Decentralized Personalized Model with Neighbor selection (DPMN) algorithm and provided a thorough experimental evaluation across a variety of datasets to demonstrate its applicability and effectiveness in federated learning environments, especially within the context of recommendation systems integrated with edge computing in cloud computing networks. Our findings not only showcase the potential of DPMN in addressing the challenges of decentralized training methods but also highlight the synergy between federated learning and edge computing as a transformative approach for privacy-preserving artificial intelligence (AI) applications.

## A. Conclusions on the Advantages of Cloud Computing and Deep Reinforcement Learning

Our research underscores the substantial benefits of leveraging cloud computing and deep reinforcement learning in enhancing the performance and efficiency of federated learning systems. The integration of cloud computing provides a scalable and flexible infrastructure capable of handling the immense computational demands of federated learning, particularly in the training and deployment of complex AI models across distributed networks. Furthermore, the application of deep reinforcement learning enables adaptive and intelligent decision-making processes within federated learning frameworks, optimizing resource allocation, and improving the overall system efficiency and model accuracy.

## B. Future Prospects

Looking ahead, the continued evolution of cloud computing technologies and advancements in deep reinforcement learning algorithms hold significant promise for further enhancing federated learning systems. The potential for developing more sophisticated and efficient model training and aggregation methods, coupled with the optimization of resource utilization in cloud environments, could lead to even greater improvements in AI model performance and system scalability. Additionally, the exploration of novel privacy-preserving techniques and security measures will be crucial in addressing the evolving challenges of data protection and regulatory compliance in federated learning applications.

## C. Advantages of Federated Learning and Edge Computing for Privacy Protection

The integration of federated learning with edge computing emerges as a particularly compelling solution for privacy protection in AI applications. By enabling model training to occur directly on edge devices without the need to transfer raw data to centralized servers, this approach significantly mitigates privacy risks and data leakage concerns. Moreover, the decentralized nature of federated learning, combined with the computational capabilities of edge computing, facilitates real-time, personalized AI applications across various domains, from healthcare and finance to smart cities and IoT, while upholding stringent privacy and data security standards.

## D. Concluding Remarks

The convergence of federated learning, edge computing, and cloud computing represents a powerful paradigm shift in the development and deployment of AI systems. Our research on the DPMN algorithm illustrates the efficacy of this integrated approach in enhancing recommendation systems, with broader implications for various AI-driven applications. As we continue to navigate the complexities of data privacy, security, and the ever-growing demand for intelligent, personalized services, the insights gained from this study will undoubtedly contribute to the advancement of more secure, efficient, and user-centric AI solutions in the cloud and edge computing era.